\documentclass[conference]{IEEEtran}
\IEEEoverridecommandlockouts
\usepackage{cite}
\usepackage{amsmath,amssymb,amsfonts}
\usepackage{algorithmic}
\usepackage{graphicx}
\usepackage{subfig}
\usepackage{multirow}
\usepackage{stfloats}
\usepackage{textcomp}
\usepackage{xcolor}
\makeatletter
\newcommand{\linebreakand}{%
  \end{@IEEEauthorhalign}
  \hfill\mbox{}\par
  \mbox{}\hfill\begin{@IEEEauthorhalign}
}
\makeatother
\def\BibTeX{{\rm B\kern-.05em{\sc i\kern-.025em b}\kern-.08em
    T\kern-.1667em\lower.7ex\hbox{E}\kern-.125emX}}

\title{Corrupting Data to Remove Deceptive Perturbation: Using Preprocessing Method to Improve System Robustness\\
{\footnotesize \textsuperscript{}}
\thanks{}
}

\author{
\IEEEauthorblockN{1\textsuperscript{st} Hieu Le}
\IEEEauthorblockA{\textit{department of Computer Science} \\
\textit{Boston University}\\
Boston, MA, USA \\
hle@bu.edu}
\and
\IEEEauthorblockN{2\textsuperscript{nd} Hans Walker}
\IEEEauthorblockA{\textit{department of Computer Science} \\
\textit{Boston University}\\
Boston, MA, USA \\
hans21@bu.edu}
\and
\IEEEauthorblockN{3\textsuperscript{th} Dung Tran}
\IEEEauthorblockA{\textit{Microsoft Research} \\
Redmond, WA \\
dung.tran@microsoft.com}
\linebreakand 
\IEEEauthorblockN{4\textsuperscript{rd} Peter Chin}
\IEEEauthorblockA{\textit{department of Computer Science} \\
\textit{Boston University}\\
Boston, MA, USA \\
spchin@bu.edu}
}
\begin{document}
\maketitle

\begin{abstract}
Although deep neural networks have achieved great performance on classification tasks, recent studies showed that well trained networks can be fooled by adding subtle noises. This paper introduces a new approach to improve neural network robustness by applying the recovery process on top of the naturally trained classifier. In this approach, images will be intentionally corrupted by some significant operator and then be recovered before passing through the classifiers. SARGAN - an extension on Generative Adversarial Networks (GAN) is capable of denoising radar signals. This paper will show that SARGAN can also recover corrupted images by removing the adversarial effects. Our results show that this approach does improve the performance of naturally trained networks. 

\end{abstract}


\section{Introduction}
The area of machine learning has been studied and researched for many decades. In recent years, the progress in computational power and the increase in data sizes and varieties have enabled a vast improvement in the performances of deep learning algorithms. This has brought forward many applications of machine learning to image recognition, speech, and natural language processing, some of which surpasses human performances \cite{Krizhevsky_2012}. Nevertheless, these well trained networks can easily be fooled by some clever changes to the input. Various works from \cite{Szegedy_2013} and \cite{Kurakin_scale} showed that imperceptible perturbations could significantly increase the error rate of a classifier. On the other hand, \cite{Nguyen_2014} found that images that do not look like anything to human eyes can fool a network classifier into mislabeling these data with high confidence. Thus, there has been active research on designing robust networks that can withstand adversarial attacks. One popular approach is adversarial training \cite{Kurakin_scale}, in which the dataset is injected with the adversarial instances so that networks can be familiarized with the attacks. On the other hand, \cite{Madry_2018} uses optimization against the "first-order" adversary approach which shows promising results on a variety of adversarial attacks.

In contrast to prior research which focuses mainly on building robust networks, this paper introduces a new method of data preprocessing to prevent adversarial attacks. Thus, instead of trying to build a network that can withstand adversarial attacks, our goal is to build a system that is capable of removing the perturbation from the data before it passes through a classifier. Specifically, we will corrupt data with noises and then recover it using a variation of generative neural networks. The motivation for this approach is a well trained denoiser network can remove extra noise and any possible adversarial perturbation, hence the target classifier may yield higher performance on these preprocessed data. In this paper, our contributions include:
\begin{itemize}
    \item Demonstrating that SARGAN - an extension of GAN - can denoise images corrupted with various operators.
    \item Using SARGAN network as a preprocessing step to clean up noisy images before passing instances to a classifier.
\end{itemize}

\section{Marterials And Method}
\subsection{Materials}
    \subsubsection{Adversarial Attacks}
    Adversarial attacks in machine learning are attacks where the main purpose is to fool an AI agent. \cite{Szegedy_2013} and \cite{Goodfellow_Shlens_2014} showed that with knowledge about the specific deep neural network classifier, it is possible to generate adversarial instances that can fool the targeted classifier. This is often referred to as \textit{white box attack}. \cite{Nguyen_2014} introduced a more sophisticated attack where the adversarial model can generate instances which could be misclassified as some specific classes. Besides \textit{white box attack}, adversarial attacks can also be achieved without any knowledge of the internal structure of the targeted neural networks \cite{Papernot_2016}. The approach that \cite{Papernot_2016} took is called \textit{black box attack}. This approach posed a real threat to machine learning in application since it does not need to know the detailed parameters of the targeted system. One could imagine such adversarial attacks may be carried out against a facial recognition security system and thus bypass that security layer. Thus, in order to apply machine learning to any critical systems, a robust defense mechanism has to be developed to minimize such threats.\\
    \subsubsection{SARGAN}
    Generative Adversarial Networks (GANs) is one of the state-of-the-art deep learning algorithms that was developed by \cite{Goodfellow_Pouget_2014}. To extend on the idea of GAN, \cite{Tran_2018} created
    SARGAN with the purpose of denoising synthetic aperture radar signals. Conventional method to recovering corrupted data such as OMP\cite{omp} and SP\cite{sp} basically solve $l_1$ or $l_0$ norm minimization problem. These methods, while proven to be sufficient, are often expensive and time consuming because of their NP-hard nature \cite{Ge_Jiang_2011}. With a twist to the generative model, SARGAN has been shown to outperform existing compressive sensing methods \cite{Tran_2018}. The key difference between SARGAN and GANs is that while GANs generative model requires random noise input \cite{Goodfellow_Pouget_2014}, the input of SARGAN is the corrupted/missing observation of some sources \cite{Tran_2018}. The basic structure of SARGAN generator is similar to the standard encoder-decoder network. SARGAN's loss function on the other hand minimizes a combination of content loss, which encourages the integrity preservation of recovered data, and adversarial loss, which encourage the generator to create more convincing instances.
    
\subsection{Method}
\label{section:method}
A common approach to adversarial defense is training neural networks with adversarial instances. This can be achieved by either including adversarial examples in the training set or creating worst-case adversarial examples online using techniques such as projected gradient descent by \cite{Madry_2018}. Each of these two approaches has its own drawback. While the former often compromises the performance on a natural uncorrupted instance, the latter is highly resource intensive and time consuming. More recently, \cite{xie2018feature} have discovered that while adversarial perturbation is small in the input space, the malicious operator is more significant in magnitude at the inner layer of the trained networks.  To enhance the robustness of adversarial training, \cite{xie2018feature} added denoising blocks at the intermediate layers. Another denoising approach is done by \cite{liao2017defense} where a denoising neural network is trained separately from a classifier. Specifically, the denoiser is an autoencoder that learns the adversarial perturbations instead of the original image. In this paper, we propose a new approach that does not require training with adversarial example but instead, we would attempt to wash away any adversarial effects on an image if exist using image preprocessing. This approach would contain the following steps:
\begin{itemize}
    \item[i)] Using SARGAN model to train a network that can recover corrupted images, specifically denoising image in this case.
    \item[ii)] Adding random noise to a given an image (natural or adversarially corrupted).
    \item[iii)] Recover the noisy image with the SARGAN model 
\end{itemize}
Our first goal is to first corrupt data with noise and build a generative model that can recover corrupted data. The corrupting method we will use is additive white Gaussian noise in which white noise will be added to each image $X = [x_{ij}]$ to gain the corrupted image $Y$. Another corrupting operator we also use to train SARGAN is mask coverage where we will cover a small patch of the image. Given a  square area with width $w$ and top left corner at $k,k$ that is relatively small to the size of the image, the operator is
\begin{equation}
Z = [z_{ij}] | z_{ij} = \begin{cases}
-x_{ij} &  k \leq i \leq (k+l)  \text{ and } k \leq j \leq (k+l) \\
0  & \text{otherwise}
\end{cases}
\end{equation}

and the corrupted image is
\begin{equation}
    Y = X + Z
\end{equation}
\subsubsection{Denoising with SARGAN}
    Given training data $\{X_i, Y_i\}$ where each $X_i$ is an image and $Y_i$ is the corrupted image, the SARGAN network \cite{Tran_2018} is trained by solving
    \begin{equation}
        argmin_{\theta_G} \sum_{i=1}^n \mathcal{L}(G_{\theta_G}(Y_i), X_i)
    \end{equation}
    where $G_{\theta_G}$ is the generative model with parameters $\theta_G$. The SARGAN model was created to reconstruct synthetic aperture data which is originally in the time domain. Thus, to feed the data into the generator network, the data have to be transformed to the frequency domain using Fourier transformation. Since our targeted data are images, we do not need to apply the transformation. The loss function is a linear combination of two terms: a content loss and an adversarial loss
    \begin{equation}
        \mathcal{L}(G_{\theta_G}(Y), X) = \ell_{\text{content}}(G_{\theta_G}(Y), X) + \lambda \ell_{\text{adversarial}}(G_{\theta_G}(Y))
    \end{equation}
    where 
    \[ \ell_{\text{content}}(G_{\theta_G}(Y), X) =  \|G_{\theta_G}(Y) -  X\|_1\]
    \[\ell_{\text{adversarial}}(G_{\theta_G}(Y)) = - \log((D_{\theta_D}((G_{\theta_G}(Y)))\]
    In $\ell_{\text{adversarial}}(G_{\theta_G}(Y))$ above, we have $D_{\theta_D}$ is the discriminator network with weights $\theta_D$.
    To evaluate the correctness of the reconstructed instance, we calculate PSNR between original and reconstructed data where PSNR is calculated as follow:
    
    \begin{equation}
        MSE =\frac{1}{m\times n} \sum_{i=0}^{m-1} \sum_{j=0}^{n-1} [X_{ij} -G_{\theta_G}(Y)_{ij}]^2
    \end{equation}
    \begin{equation}
        PSNR = 10 \times \log_{10}(\frac{(\max X_{ij})^2}{MSE})
    \end{equation}\\
    \subsubsection{Adversarial Attack with Projected Gradient Descent (PGD)}
    The method that we use to create adversarial examples is the projected gradient descent attack. The idea is very much similar to the gradient descent approach in training a neural network. During training, we want to minimize the loss value with respect to the pair of network output and label. In PGD attack, we create the adversarial instances by going the opposite direction:
    \begin{equation}
        argmax_{\delta \in \Delta} \mathcal{L}(F_{\theta_F}(X + \delta), Y)
    \end{equation}
    Thus, given a network $F$, we want to find a perturbation $\delta$ that maximizes the loss for a given output. Ideally, we can bound $\delta$ to some small value so that the noise is unnoticeable to human eyes. It is also worth noting that a new $\delta$ needs to be calculated for each image.

\section{Experiments}

The datasets that we use in this paper are the CIFAR-10 and CIFAR-100 datasets by \cite{cifar10}, MNIST dataset by \cite{mnist} , and fashion-MNIST dataset by \cite{xiao2017fashionmnist}. \\

\subsection{Denoising images with SARGAN}
\label{sec3.1}
For MNIST dataset, given that the image data is in the range $0,1$, we trained four SARGAN models with the input images having white Gaussian noise with a standard deviation between $0.0, 0.5$ added and the other with a mask coverage of size $4$ by $4$ to $6$ by $6$. The position of the mask is also picked randomly within a few pixels from the image center.

\begin{figure}[h]
\centering
  \subfloat[]{\includegraphics[width=5.5cm]{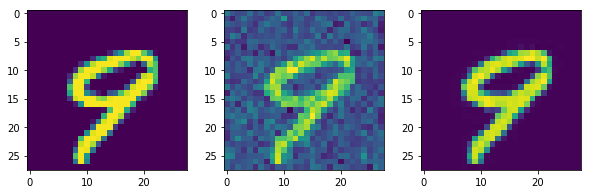}\label{fig1}}
  \hspace{0.5cm}
  \subfloat[]{\includegraphics[width=5.5cm]{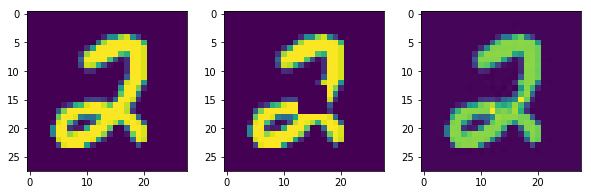}\label{fig1}}
  \captionof{figure}{SARGAN recovery on MNIST data corrupted by Gaussian noise $(a)$ and mask coverage $(b)$. On the Left column:  original images, Center column: corrupted images, Right column: recovered images.}
\end{figure}

The trained SARGAN is then tested against 100 images and sample results can be seen in Figure 1 (a) and (b). We also calculate the peak signal-to-noise ratio (PSNR) values between original and reconstructed images. The average PSNR values are 26.2 for Gaussian noise recovery and 28.7 for mask coverage recovery. The process is similar for the Fashion-MNIST dataset.

\begin{figure}[h]
\centering
\includegraphics[width=7cm]{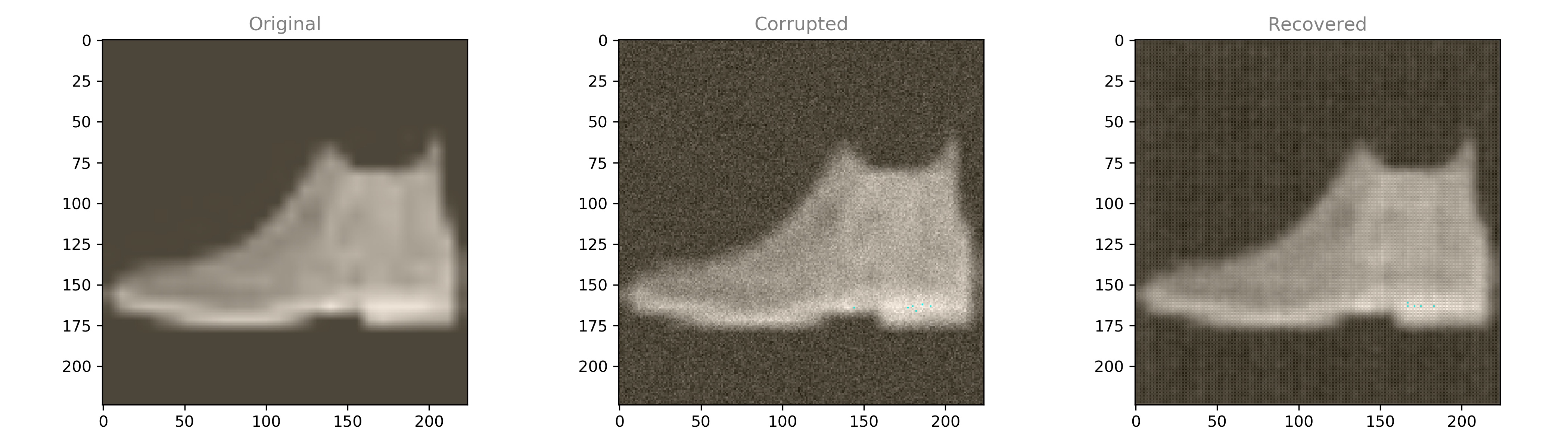}

  \caption{SARGAN recovery on FashionMNIST images corrupted by Gaussian noise.}
\end{figure}

For the CIFAR-10 and CIFAR-100 datasets, SARGAN is trained with $50,000$ instances which are corrupted by white Gaussian noise with a standard deviation between 0 and 0.12. We chose a smaller white noise because CIFAR-10 images are larger and contain more complicated features than MNIST images. 
\begin{figure}[htbp]
\centering
 \subfloat[]{\includegraphics[width=5.5cm]{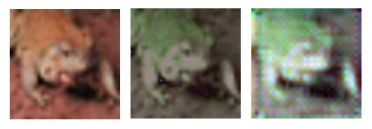}\label{fig3}}\\
 \subfloat[]{\includegraphics[width=5.5cm]{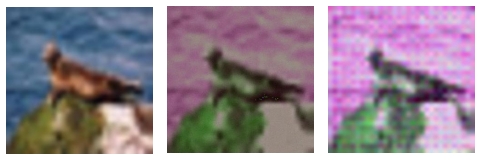}\label{fig3}}
\captionof{figure}{SARGAN recovery on CIFAR-10 $(a)$ and CIFAR-100 $(b)$ images corrupted by Gaussian noise}
\end{figure}
After training, SARGAN network then is tested against $10,000$ new images. The average PSNR value of the test images is 26.4 for CIFAR-10 and 18.83 for CIFAR-100. From testing on two datasets with two corrupting operators, we see that SARGAN can reconstruct images reasonably.\\

\begin{figure}[!htbp]
\centering
\includegraphics[width=0.5\textwidth]{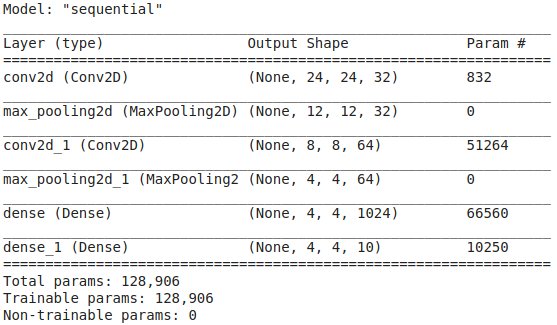}
\captionof{figure}{Classifier network architecture for MNIST, Fashion-MNIST and CIFAR-10 datasets}
\label{fig:target_model_arch}
\end{figure}

\subsection{Preprocessing adversarial instances}
The SARGAN network trained in section $3.1$ is then used to preprocess the images as described in step $(ii)$ of Section \ref{section:method}. To experiment on the adversary-removing ability of the SARGAN denoisers, we first need to construct the adversarial instances. We use the PGD attack with cross-entropy loss on the targetted model described in the section below. With each dataset, we estimate the smallest possible adversarial parameter $\varepsilon$ described by \cite{Madry_2018} where the accuracy of the model classifying adversarial instances approaches $0 \%$ and use those $\varepsilon$ to create adversarial instances. The $\varepsilon$ we used for MNIST, Fashion-MNIST, CIFAR-10 and CIFAR-100 are $0.15, 0.08, 0.01$ and $0.005$ accordingly. For each dataset, we use $10,000$ images for testing. These adversarial instances are then corrupted using additive white Gaussian noise and reconstructed by SARGAN.

In terms of the classification task, for the MNIST and Fashion-MNIST datasets, we build a simple architecture with 2 convolutional layers of 32 and 64 kernels, each followed by a maxpooling layers followed by a dense layer. The detailed architecture is shown in Figure \ref{fig:target_model_arch}. For the CIFAR-10  and CIFAR-100 datasets, we use transfer learning with a ResNet-50 model trained on Imagenet and adapt it to the two datasets.

\begin{table*}[!htbp]
\caption{Accuracy of classifiers trained with natural images testing on natural images, natural images with adversarial perturbation (adversaries), adversaries $+$ Gaussian noise passing through 1 SARGAN, 2 SARGANS, 3 SARGANS and 4 SARGANS}
\begin{center}
\begin{tabular}{|c|c|c|c|c|c|c|}
\hline
 \multirow{2}{*} & Nat. Imgs & Adv. Imgs & Adv. Imgs  & Adv. Imgs & Adv. Imgs & Adv. Imgs \\
 &&&1 SARGAN & 2 SARGANs & 3 SARGANs & 4 SARGANs \\
\hline
{MNIST} & 99.12\% & 0.66\% & 46.9\% & 88.96\% & 89.1\% & 89.08\%\\
\hline
{Fashion-MNIST} & 90.66\% &  0\% & 21.61\%  & 53.22\% & 70.51\% & 71.76\% \\
\hline
{CIFAR-10} & 89.11\% &  0.99\% & 21.15\% & 45.83\% & 45.89\% & 51.58\% \\
\hline
{CIFAR-100} & 60.83\% &  0.69\% & 24.23 \% & 23.86\%& 24.26\% & 29.29\%\\
\hline
\end{tabular}
\end{center}
\label{tab:accuracy_multi_sargan}
\end{table*}

\subsection{Filterting adversarial instances with multiple SARGAN denoisers}
As we experimented with using a SARGAN autoencoder-decoder as a denoiser, we want to explore the idea of using multiple SARGAN denoisers to test the performance against a single denoiser. In this experiment, we pass the images with additive Gaussian noise through some number of denoisers (from one to four) to see if adding more denoisers would help increase the adversarial robustness of the system. We train the SARGAN denoisers sequentially by using the output of the previous denoiser as training input for the next denoiser and use the clean original image as the training target. For example, denoiser $2$ would take the output of denoiser $1$ as input and the corresponding clean image from the dataset as the target output. The first SARGAN denoiser is trained as described in section \ref{sec3.1}. 
\begin{figure*}[h]
\centering
\includegraphics[width=12cm]{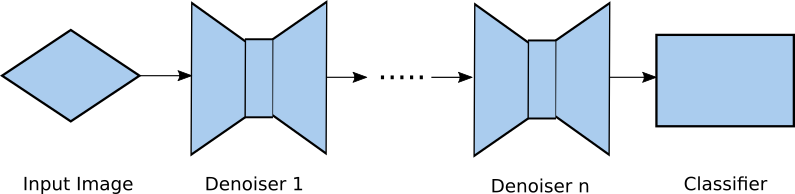}
\caption{Preprocessing images with multiple SARGAN denoisers scheme}
\end{figure*}

\subsection{Robustness of Denoiser against adversarial instances}
\label{robustness_of_denoiser_against_adv_instance}
Lastly, we test the robustness of SARGAN denoisers with respect to adversarial perturbation magnitude. With our denoisers trained with Gaussian noise, we gradually increase the perturbation limit $\varepsilon$ and observe the effectivenese of preprocessing images with SARGAN compared with no preprocessing.

\section{Discussion}

\subsection{Classification performance}
With the result from using series of SARGAN Gaussian denoiser as a preprocessing method, we showed that by intentionally adding Gaussian noise to adversarial images and passing them through the denoisers, it is possible to wash away a significant level of adversarial perturbation. As shown in Table \ref{tab:accuracy_multi_sargan}, using four SARGANs yields the best result in terms of classification accuracy for model trained with natural images and tested on adversarial images across all four datasets. For all datasets, the naturally trained classifier yield near $0\%$ accuracy when given adversarial instances generated from the test sets. After passing through a series of four SARGANs, the accuracy on these instances improved to $89.08\%, 71.76\%, 51.58\%$ and $29.29\%$ for MNIST, Fashion-MNIST, CIFAR-10 and CIFAR-100 accordingly.

\begin{table}[tbp]
\caption{Comparison of accuracy of classifiers trained on regular images vs PGD trained for 4 datasets. The comparisons includes regular images, adversarial images and regular/adversarial images with additional Gaussian noise and processed through 4 SARGAN denoisers.}
\begin{center}
\resizebox{\columnwidth}{!}{%
\begin{tabular}{|c|c|c|c|}
\hline
  & & Nat. Trained & PGD Trained \\
\hline
\multirow{4}{*}{MNIST} & {Reg. Images} & 99.12\% & 98.41\% \\
                       & {Reg. Images + 4 SARGANs} & 98.07\% & 97.05\% \\
                       & {Adv. Images} & 0.66\% & 95.55\% \\
                       &  {Adv. Images + 4 SARGANs} & 89.08\% & 93.01\%\\
\hline
\multirow{3}{*}{F-MNIST} & {Reg. Images} & 90.66\% & 76.88\% \\
                       & {Reg. Images + 4 SARGANs} & 85.4\% & 76.11\% \\
                       & {Adv. Images} & 0\% & 67.43\% \\
                       &  {Adv. Images + 4 SARGANs} & 71.76\% & 70.06\%\\
\hline
\multirow{3}{*}{CIFAR-10} & {Reg. Images} & 89.11\% & 88.63\% \\
                       & {Adv. Images} & 0.99\% & 34.43\% \\
                       &  {Adv. Images + 4 SARGANs} & 51.58\% & 58.94\%\\
\hline
\multirow{3}{*}{CIFAR-100} & {Reg. Images} & 60.83\% & 61.08\% \\
                       & {Adv. Images} & 0.69\% & 12.03\% \\
                       &  {Adv. Images + 4 SARGANs} & 29.29\% & 36.5\%\\
\hline
\end{tabular}%
}
\end{center}
\label{tab:comparison_adv_trained_vs_nat_trained}
\end{table}

In addition, we also evaluate our approach against PGD trained method described by \cite{Madry_2018}. We compare the performance of a regularly trained classifier with a PGD trained with respect to regular images, adversarial images generated by PGD attack and regular/adversarial images with our preprocessing approach. From table \ref{tab:comparison_adv_trained_vs_nat_trained}, we can observe that with Fashion-MNIST, CIFAR-10 and CIFAR-100, using a SARGAN preprocessing approach together with a naturally trained classifier yields better results than PGD trained classifiers without any preprocessing step. Also, SARGAN preprocessing helps increasing the accuracy for both natural and PGD trained classifier. On MNIST and Fashion-MNIST, applying SARGAN preprocessing on regular images only reduce accuracy by a small percentage point for both regular and PGD trained network, the reduction in accuracy are mostly within $1$ percentage point except for Fashion-MNIST naturally trained classifier at $4 \%$ point accuracy reduction. Thus, experiments showed that the benefits of removing adversarial effects significantly outweigh the cost of accuracy for both regular and adversarial images.

\subsection{Robustness against adversarial instances}
\begin{figure}[!hbtp]
\centering
  \subfloat[MNIST ]{\includegraphics[width=6cm]{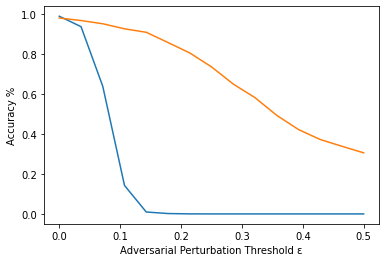}\label{fig1}}
 \hspace{0.1cm}
  \subfloat[FashionMNIST ]{\includegraphics[width=6cm]{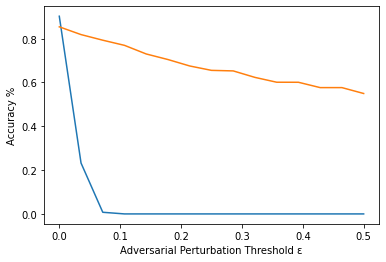}\label{fig2}}
\hspace{0.1cm}
  \caption{Accuracy on classifier trained on clean images and tested on adversarial images of MNIST and Fashion-MNIST. In Blue: adversarial images without any preprocessing. In Orange: adversarial images with Gaussian noise added and preprocessed with 4 SARGAN denoisers.}
 \label{fig:mnist_fmnist_robust_epsilon}
\end{figure}

\begin{figure}[htbp]
\centering
  \subfloat[CIFAR-10 ]{\includegraphics[width=6cm]{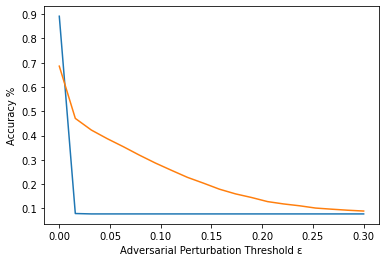}\label{fig2}}
\hspace{0.1cm}
  \subfloat[CIFAR-100 ]{\includegraphics[width=6cm]{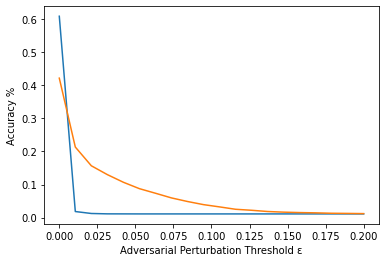}\label{fig2}}
  \caption{Accuracy on classifier trained on clean images and tested on adversarial images of CIFAR-10 and CIFAR-100. In Blue: adversarial images without any preprocessing. In Orange: adversarial images with Gaussian noise added and preprocessed with 4 SARGAN denoisers.}
\label{fig:cifar10_cifar100_robust_epsilon}
\end{figure}

Beside evaluating the accuracy of our preprocessing method against PGD training, we test the robustness of our model against stronger adversarial instances as describe in Section \ref{robustness_of_denoiser_against_adv_instance}. Experimental results in Figure \ref{fig:mnist_fmnist_robust_epsilon} and Figure \ref{fig:cifar10_cifar100_robust_epsilon} showed that SARGAN preprocessing is qute robust against perturbation increase for MNIST and Fashion-MNIST datasets. With harder datasets like CIFAR-10 and CIFAR-100, preprocessing adversarial instances with denoisers also flatten the curve and delay the accuracy deficiency much better than without using the denoisers. The overall observable trend is that while the accuracy of a regular classifier drop to near-zero very quickly as we use any significant value of $\varepsilon$, the same classifier with our prepocessing approach can withstand a stronger adversarial attacks longer before the performance drops to the same low level as without a preprocessing step. 

While the adding the preprocessing step does not make the performance of a naturally trained classifer better than a PGD trained network, the SARGAN recovery method has shown its ability in flattening the adversarial effect. More importantly, while this work also utilizes autoencoders like \cite{liao2017defense}, the key difference and advantage of this preprocessing approach is that unlike adversarial training and denoising in \cite{liao2017defense} and \cite{xie2018feature} where a specific type of corrupted images is included in the training, the preprocessing approach does not make any assumption on the type of adversarial attacks. Thus, it is highly generic and can be applied to any form of low noise adversarial attack, regardless of attacks based on any $L^p$ norm. We believe that while the idea of adding white noise on top of the input is counter-intuitive at first glance, it is novel. Instead of focusing on building a neural network that can withstand adversarial attacks, we instead ask the question of how to remove the adversarial effects from data. This method can be thought of as a filter to remove the malicious operator. Since many adversarial attack methods like \cite{Kurakin_scale} and \cite{Szegedy_2013} are very subtle numerically and visually, by first corrupting data with noises, the adversarial effects may be diluted or even dominated by the random white noise. Thus, a well trained generator can reconstruct the natural instances from this corrupted instance. 

\section{Conclusion}
Our results showed that while the performance of SARGAN recovery method is not yet at the same level with state of the art robust trained model, this method provides a new angle to designing robust systems. Specifically, it can be used as a filtering step before passing an instance to the model and will improve accuracy of the model whether the classifier is trained with regular images or adversarial images. In addition, we also shows that a series of SARGAN denoiser trained in the same fashion is more effective in washing away the adversarial effect than one single SARGAN denoiser. Beyond this work, we plan to apply this method to other forms of adversarial attacks as well as optimize the model to better filtering adversarial effects. If we could increase the classification performance of our adversarial filtering approach to the same level with state of the art adversarial defense, it will provide an alternative approach to adversarial defense as our approach separates the adversarial robustness from direct classification model training and instead looks at it as an adversarial filtering problem, which allow researchers to study and focus on the classification and adversarial filtering tasks separately.


\bibliographystyle{IEEEtran}  
\bibliography{citation}

\end{document}